\documentclass[sigplan,nonacm,screen]{acmart}

\usepackage{amsmath}
\usepackage{graphicx} 
\usepackage{caption}  
\usepackage{subcaption} 

\makeatletter
\renewcommand\p@subfigure{\thefigure-}
\makeatother
\usepackage{booktabs}
\usepackage{geometry}
\usepackage{tikz}
\usepackage{siunitx}
\usepackage{multirow}
\usepackage{makecell}
\usepackage[table]{xcolor}

\begin{document}

\title{Energy-Efficient and Dequantization-Free Q-LLMs: A Spiking Neural Network Approach to Salient Value Mitigation}

\author{Chenyu Wang}

\affiliation{%
  \institution{Sun Yat-sen University}
  \city{Guangzhou}
  \state{Guangdong}
  \country{China}
}
\email{wangcy236@mail2.sysu.edu.cn}

\author{Zhanglu Yan}
\authornote{Corresponding authors are Zhi Zhou and Zhanglu Yan.}

\affiliation{%
  \institution{National University of Singapore}
  \country{Singapore}}
\email{e0385109@u.nus.edu}

\author{Zhi Zhou}
\authornotemark[1]
\affiliation{%
  \institution{Sun Yat-sen University}
  \city{Guangzhou}
  \state{Guangdong}
  \country{China}
}
\email{zhouzhi9@mail.sysu.edu.cn}

\author{Xu Chen}
\affiliation{%
 \institution{Sun Yat-sen University}
  \city{Guangzhou}
  \state{Guangdong}
  \country{China}
}
\email{chenxu35@mail.sysu.edu.cn}

\author{Weng-Fai Wong}
\affiliation{%
  \institution{National University of Singapore}
  \country{Singapore}}
\email{wongwf@nus.edu.sg}


\newcommand{\TODO}[1]{{\textcolor{blue}{#1}}}

\begin{abstract}

In the era of large language models (LLMs), weight-activation quantization helps fit models on edge device by reducing memory and compute bit-widths. However, three challenges persist for energy constrained hardware: (1) even after quantization, multiply-accumulate (MAC) operations remain unavoidable and continue to dominate energy consumption; (2) dequantization (or per-tensor/ channel rescaling) introduces extra arithmetic and data movement, increasing latency and energy; (3) uniform parameters bit widths clip salient values—while intra-channel mixed precision is generally impractical on current matrix hardware and memory. In contrast, brain-inspired Spiking Neural Networks (SNNs), owing to their binary spike-based information representation and the Integrate-and-Fire (IF) paradigm, naturally support mixed-precision storage and energy-efficient computation by replacing complex MACs with temporal Accumulate (ACCs). Motivated by this property, we propose SpikeQuant, which selectively applies mixed-precision quantization to activations with salient values and re-encodes them into binary spike counts, thereby enabling dynamic mixed storage of different bitwidths. Furthermore, by embedding the quantization scale into the threshold of the IF mechanism, our approach performs energy-efficient linear transformations on weights and activations while avoiding explicit dequantization. Experimental results demonstrate that SpikeQuant consistently achieves near-FP16 perplexity under W4A4 quantization while reducing energy cost by up to 4.6× compared to existing methods, highlighting its effectiveness for accurate and energy-efficient LLM deployment. 


\end{abstract}





\maketitle

\begin{figure}[h]
    \centering

    \begin{minipage}[t]{0.48\columnwidth}
        \centering
        \includegraphics[width=\columnwidth, keepaspectratio]{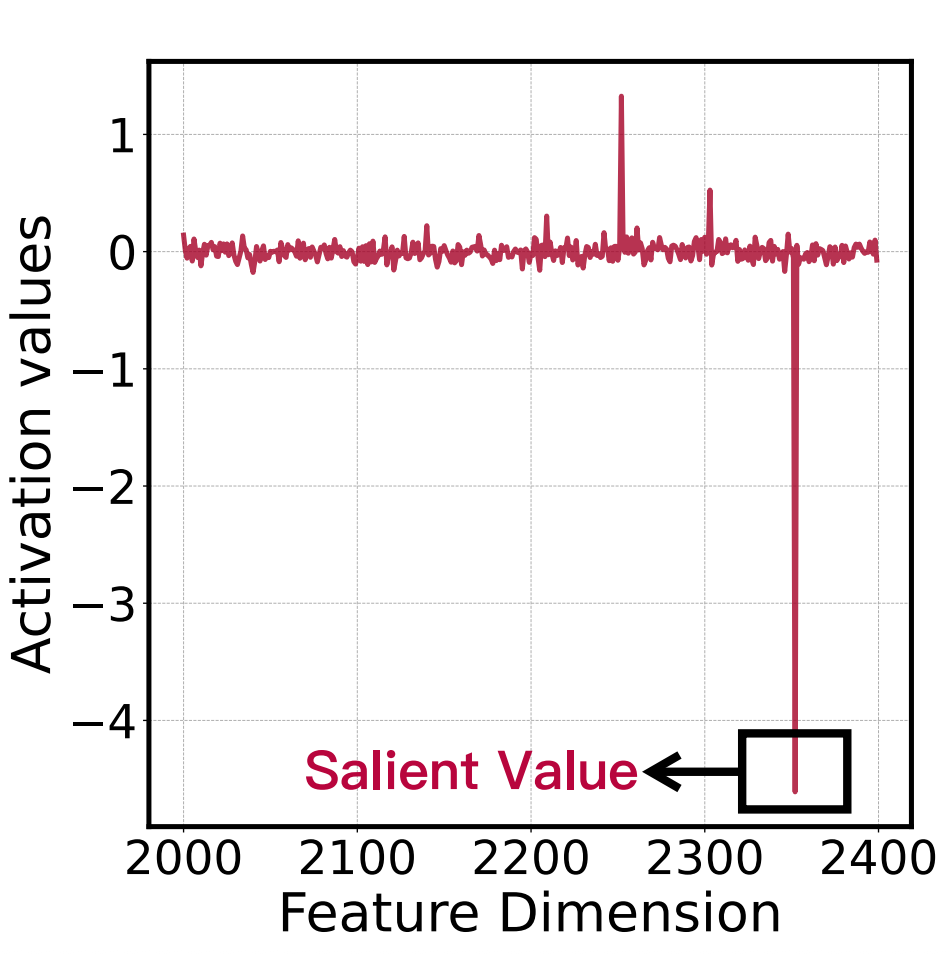}
        \captionof{figure}{Activations distribution of Llama2-7B during inference}
        \label{fig1}
    \end{minipage}
    \hfill
    \begin{minipage}[t]{0.48\columnwidth}
        \centering
        \includegraphics[width=\columnwidth, keepaspectratio]{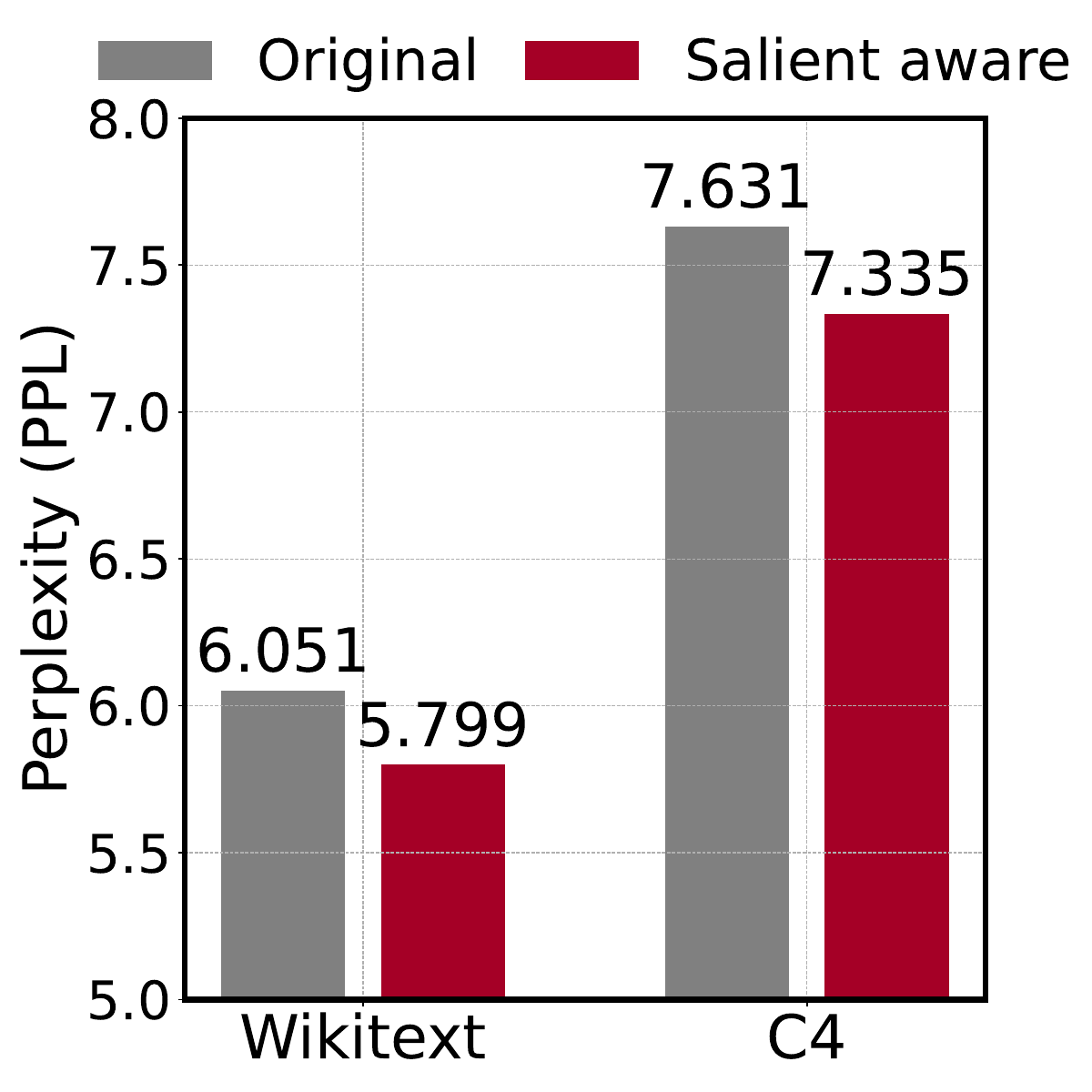}
        \captionof{figure}{PPL of original and salient aware quantization of Llama2-7B}
        \label{fig2}
    \end{minipage}

\end{figure}

\section{Introduction}
In recent years, Artificial Neural Networks (ANN) based LLMs with billions of parameters \cite{touvron2023llama} \cite{jiang2023mistral7b} have greatly advanced applications in browser inference engines \cite{ruan2024webllm} and web enhanced question-answering systems \cite{liu2023webglm}. Concurrently, there has been a growing demand for deploying such models in Web-of-Things (WoT) environments, driven by privacy, latency considerations for devices in mobile WoT settings. However, memory and energy budgets remain the bottleneck—for instance, FP32 Llama2-7B alone stores ~28 GB of weights and a single NVIDIA A6000 draws ~72 W over 10 s of inference \cite{poddar-etal-2025-towards}. To address these issues, weight-activation quantization converts full-precision parameters into low-bit representations (i.e., INT8 or INT4), thereby reducing the memory footprint and accelerating inference \cite{li2024transformerlitehighefficiencydeploymentlarge}.

However, accurate quantization LLMs are hard: activations frequently contain salient outliers that, under low bit-widths, set an overly wide scale, compressing typical values and hurting accuracy, as shown in Fig. \ref{fig1} and Fig. \ref{fig2}. Such issue is typically more severe than for weights, whose post-training distributions are near-Gaussian and relatively stable \cite{an2025systematic,xiao2023SmoothQuant}. Atom\cite{MLSYS2024_5edb57c0} and AWQ \cite{linAWQ2024} mitigate salient values via salient channel reordering or per-channel rescaling. However, they rely on time consuming offline calibration and search processes, and their fine-grained scaling strategies (e.g., intra-channel or group-wise) are expensive on current hardware, leading to tensor fragmentation, complicated scheduling and higher energy consumption.

Compared to ANN-based LLMs, the human brain, with roughly 86 billion neurons, achieves far greater efficiency. Based on this inspiration, Spiking Neural Networks (SNNs) \cite{schuman2022opportunities}, which leverage binary spike-based information and the Integrate-and-Fire (IF) paradigm, have emerged as a promising approach for next-generation machine intelligence. The advantage of SNNs lies in the full utilization of the temporal dimension, where parameters are transmitted as 0/1 spikes across \(T\) steps, and the conversion between different quantization bits only requires changing \(T\). When applied to the quantization of ANNs, this allows for (1) performing in-place mixed quantization of salient and normal values without channel reorder or channel rescaling, and (2) replacing Multiply-Accumulate (MAC) operations with energy-efficient spike Accumulations (Acc), which reduces the computation energy \cite{guo2023direct}. 

Despite their advantages for quantizing salient values, the application of SNNs remains limited to smaller models (e.g., VGG, ResNet, and BERT) \cite{wei2024q, wei2025qpsnn, liNeuronQuant2025, balexploring2024}. This limitation primarily stems from the substantial memory and data movement costs of rate encoding (where a parameter's value corresponds to its spike count over $T$ time steps) and the information loss incurred when the residual membrane potential is discarded in standard IF neurons from improper thresholding.

In terms of parameter representation, we address the salient value phenomenon—which is more pronounced in activations and more challenging to handle due to their real-time generation during inference—by encoding activations into a time-to-first-spike (TTFS) scheme \cite{bonilla2022analyzing}, where information is represented solely by the latency of the first spike. Compared with the rate encoding commonly used in SNNs, TTFS requires only a single spike within a time window, thereby significantly reducing data movement costs and memory footprint. Regarding the IF mechanism, we define the firing threshold as a function of the scales in mixed-precision quantization, such that the same mathematical effect as dequantization can be achieved without explicitly performing the dequantization process. As far as we know, in the ANN domain, this is the first work to leverage TTFS-based SNNs for mitigating salient values, improving model accuracy while reducing inference energy consumption; in the SNN domain, we bridge the gap by introducing a scale-based threshold, enabling SNN-LLMs to achieve accuracy comparable to, or even on par with, their ANN counterparts.

Our main contributions are summarized as follows:

\begin{itemize}
	\item We propose SpikeQuant, the first framework that leverages TTFS encoding in quantized LLMs, effectively  mitigates the impact of salient values.
	\item We embed mixed-precision quantization scales into the fire threshold in SNN, ensuring mathematical equivalence to dequantized linear transformations without explicitly performing dequantization.
    \item We conduct extensive motivation studies and experiments, demonstrating that SpikeQuant achieves near-FP16 perplexity and up to 78.5\% energy savings compared to state-of-the-art baselines, and narrows the performance gap between SNN-LLMs and ANN-LLMs.
\end{itemize}

\section{Preliminaries}

\subsection{Low-bit Quantization}

Low-Bit quantization maps a real-valued tensor \(x\) to an integer tensor \(x_{\mathrm{int}}\) via a scale \(S>0\) and a zero-point \(Z\in\mathbb{Z}\). The asymmetric quantization and dequantization steps are typically defined as:

\begin{equation}
    x_{\mathrm{int}} = \mathrm{clamp}\!\left(\mathrm{round}\Bigl(\frac{x}{S}\Bigr) + Z,\; q_{\min},\; q_{\max}\right),
\end{equation}

\begin{equation}
\hat x = S\,(x_{\mathrm{int}} - Z),
\end{equation}
where \(q_{\min}, q_{\max}\) denote the limits of the integer range (e.g. for unsigned 8-bit, \([0,2^{8}-1]\)), and clamp ensures values stay in that range. In contrast, symmetric quantization sets the zero-point \(Z = 0\), and chooses the scale based on the maximum of the absolute value

\begin{equation}
S = \frac{\max\bigl(|r_{\min}|, |r_{\max}|\bigr)}{q_{\max}},
\end{equation}
where \([r_{\min}, r_{\max}]\) is the clipping range of the floating values. 

Asymmetric quantization, by incorporating a zero-point offset, more accurately represents skewed or non-symmetric data distributions, thereby minimizing quantization error compared to the rigid, zero-centered range of symmetric methods. Therefore, the quantization method selected in this paper is asymmetric quantization

\subsection{Spiking Neural Networks}

We firstly focus on the IF paradigm in SNN: a neuron maintains a membrane potential \(V(t)\) that integrates incoming inputs over discrete time steps. We denote \(I(t)\) as the synaptic input at time time \(t\), and the neuron emits a spike \(s(t)\) if \(V(t)\) exceeds or equals a threshold \(V_{\mathrm{th}}\), and otherwise remains silent:

\begin{equation}
V(t) = V(t-1)+I(t) - s(t) \cdot V_{\mathrm{th}}, 
\end{equation}

\begin{equation}
s(t) = 
\begin{cases}
0, & v(t) < V_{\mathrm{th}},\\
1, & v(t) \ge V_{\mathrm{th}}.
\end{cases}
\end{equation}

There are several ways to encode input or activation values using spike trains \cite{kim2022rateencoding, guo2021neural}. The most common is rate encoding, where information is represented by the number (or frequency) of spikes over a time window $T$. Although simple, this incurs a large number of spikes, increasing data movement, and memory cost. A more efficient alternative is TTFS encoding, where each neuron is allowed at most one spike per window $T$. The timing, or latency, of that first spike encodes the magnitude (stronger or more urgent inputs spike earlier). TTFS reduces the total number of spikes and can dramatically lower memory access overhead compared to rate encoding.

\section{Motivation Experiments}

This section examines the influence of quantization bitwidth on both activations and weights, as well as the distribution of activations during inference. Building on these observations, we further discuss the challenges that arise when implementing mixed-precision quantization.

\subsection{Importance of Activations}

\begin{figure}[h]
\centering  

\begin{subfigure}[t]{0.49\linewidth} 
    \centering 
    \includegraphics[width=4.6cm]{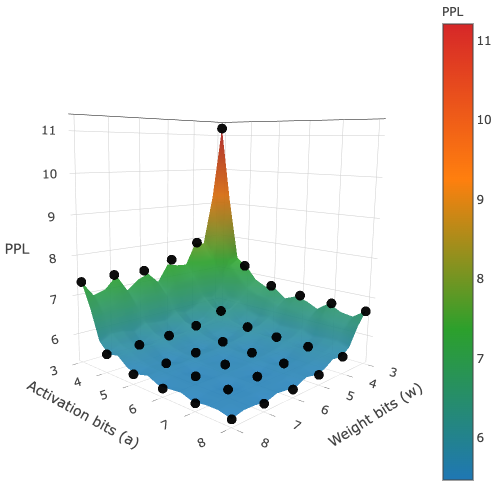}
    \caption{Wikitext-2}
    \label{motivation1.sub.1}
\end{subfigure}
\hfill 
\begin{subfigure}[t]{0.49\linewidth} 
    \centering 
    \includegraphics[width=4.6cm]{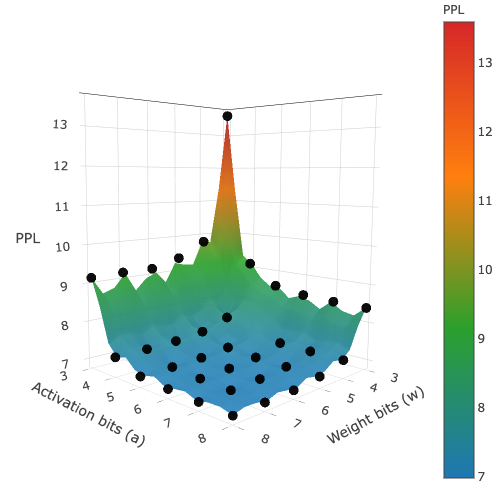}
    \caption{C4}
    \label{motivation1.sub.2}
\end{subfigure}

\caption{Perplexity of Llama-2-7B across different weight--activation quantization bitwidths}
\label{motivation1}
\end{figure}

Weight–activation quantization has emerged as a prominent strategy for efficient model compression\cite{lee2023enhancing}. To investigate the respective impacts of quantization bitwidth on weights and activations, we evaluate Llama2-7B under 1–8 bit quantization on the Wikitext-2 \cite{merity2017pointer} and C4 \cite{raffel2020exploring} datasets, with perplexity (PPL) as the metric. The results are presented in Fig. \ref{motivation1}.

Under extremely low-bit quantization settings (e.g., 1-bit and 2-bit), LLMs suffer from severe information loss, leading to a collapse in performance, as indicated by perplexity values rising to the order of $10^5$. Therefore, we report only the results for 3–8 bit quantization. As shown in Fig. \ref{motivation1}, increasing the activation bitwidth yields a more pronounced reduction in perplexity compared to increasing the weight bitwidth. In particular, the 4w4a configuration (4-bit weights and 4-bit activations) demonstrates an effective balance between efficiency and performance. However, further increasing the bitwidth results in diminishing improvements due to marginal effects, thereby limiting the potential to achieve an optimal trade-off between memory footprint and model accuracy.

Based on the above experimental conclusions and the importance of activations, we quantize both weights and activations to 4-bit, while quantizing the activation salient values to 5-bit. It is worth noting that in prior work the activation salient values are typically quantized to 8-bits or higher \cite{Bondarenko2023Quantizable,MLSYS2024_5edb57c0}, but such a scheme often incurs memory overhead that is not well-justified by the performance gains.

\subsection{Distribution of Activations}


\begin{figure}[h]
\centering  

\begin{subfigure}[t]{0.49\linewidth}
    \hspace{-0.25cm}\includegraphics[width=4.5cm]{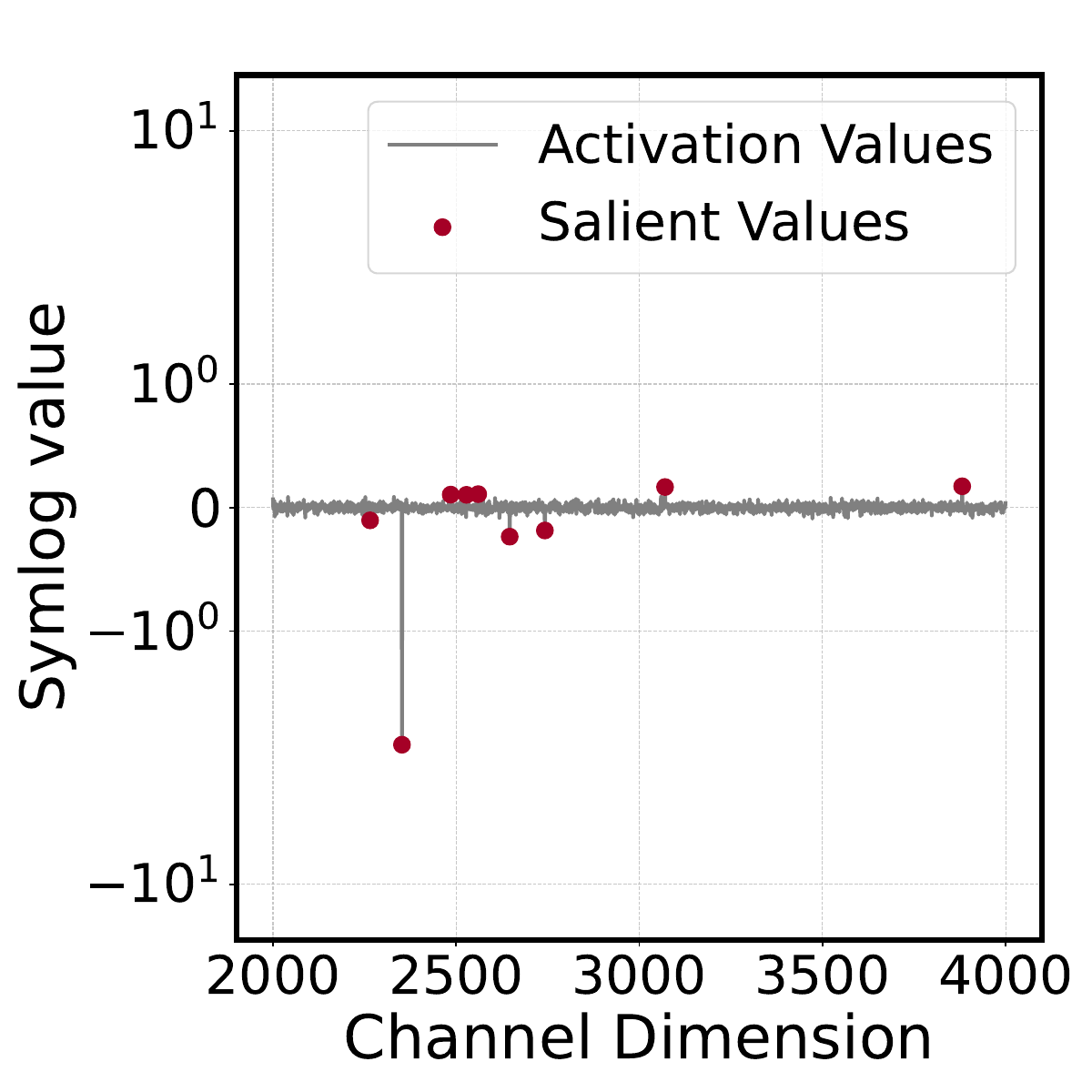}
    \caption{$o_{proj}$ of layer 26}
    \label{motivation2.sub.1}
\end{subfigure}
\hfill 
\begin{subfigure}[t]{0.49\linewidth}
    \hspace{-0.25cm}\includegraphics[width=4.5cm]{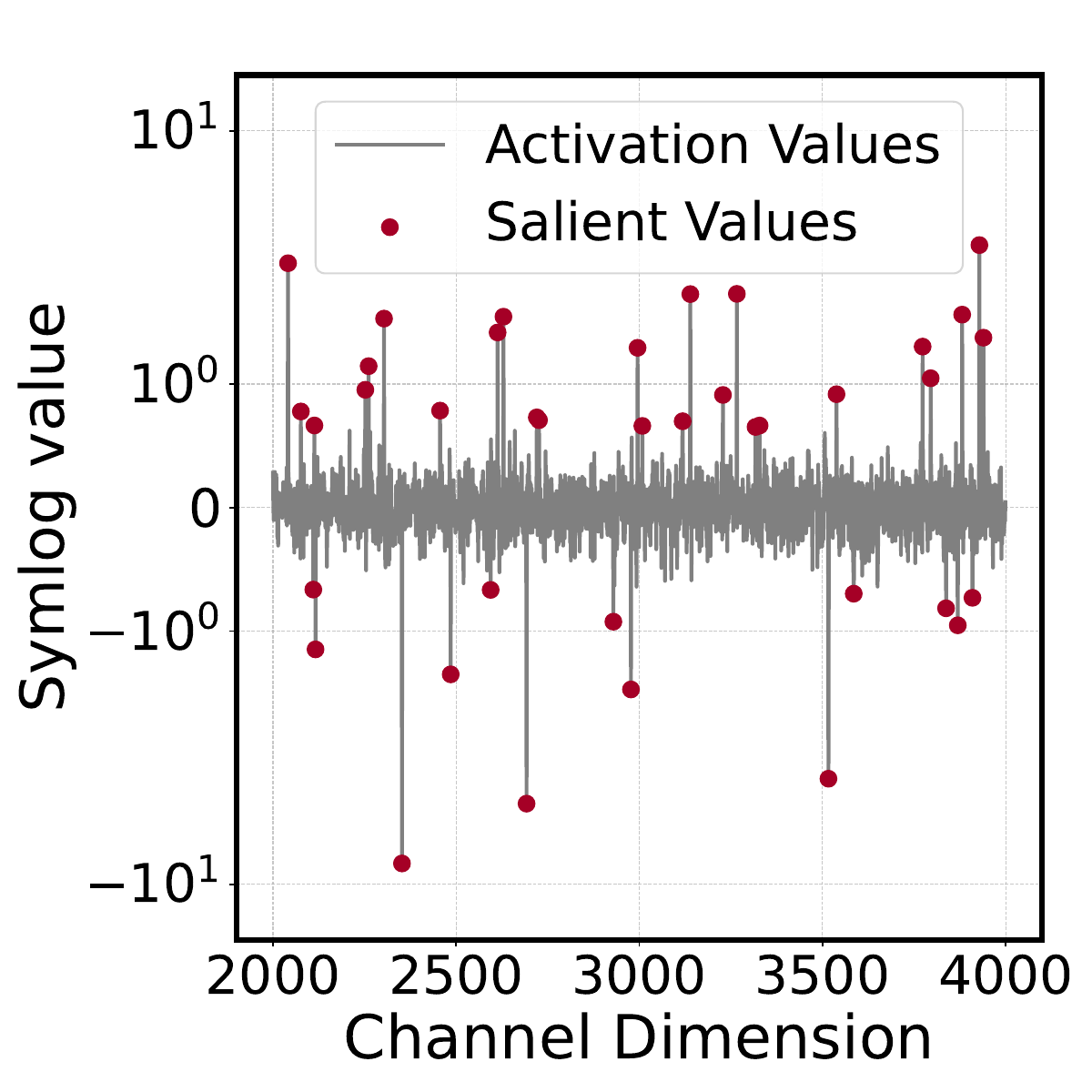}

    \caption{$o_{proj}$ of layer 31}
    \label{motivation2.sub.2}
\end{subfigure}

\begin{subfigure}[t]{0.49\linewidth}
     \hspace{-0.25cm}\includegraphics[width=4.5cm]{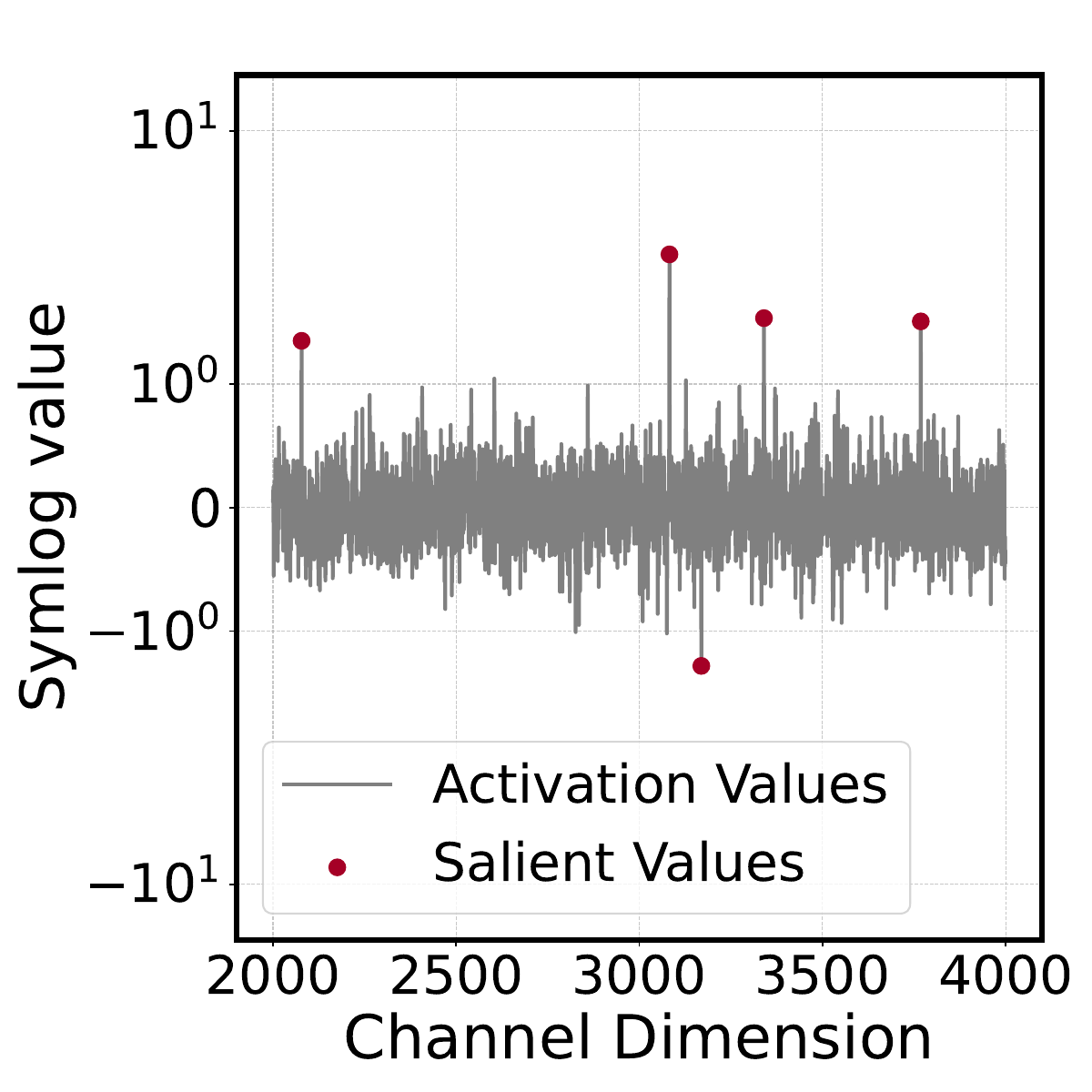}
    \caption{$up_{proj}$ of layer 26}
    \label{motivation2.sub.3}
\end{subfigure}
\hfill 
\begin{subfigure}[t]{0.49\linewidth}
    \hspace{-0.25cm}\includegraphics[width=4.5cm]{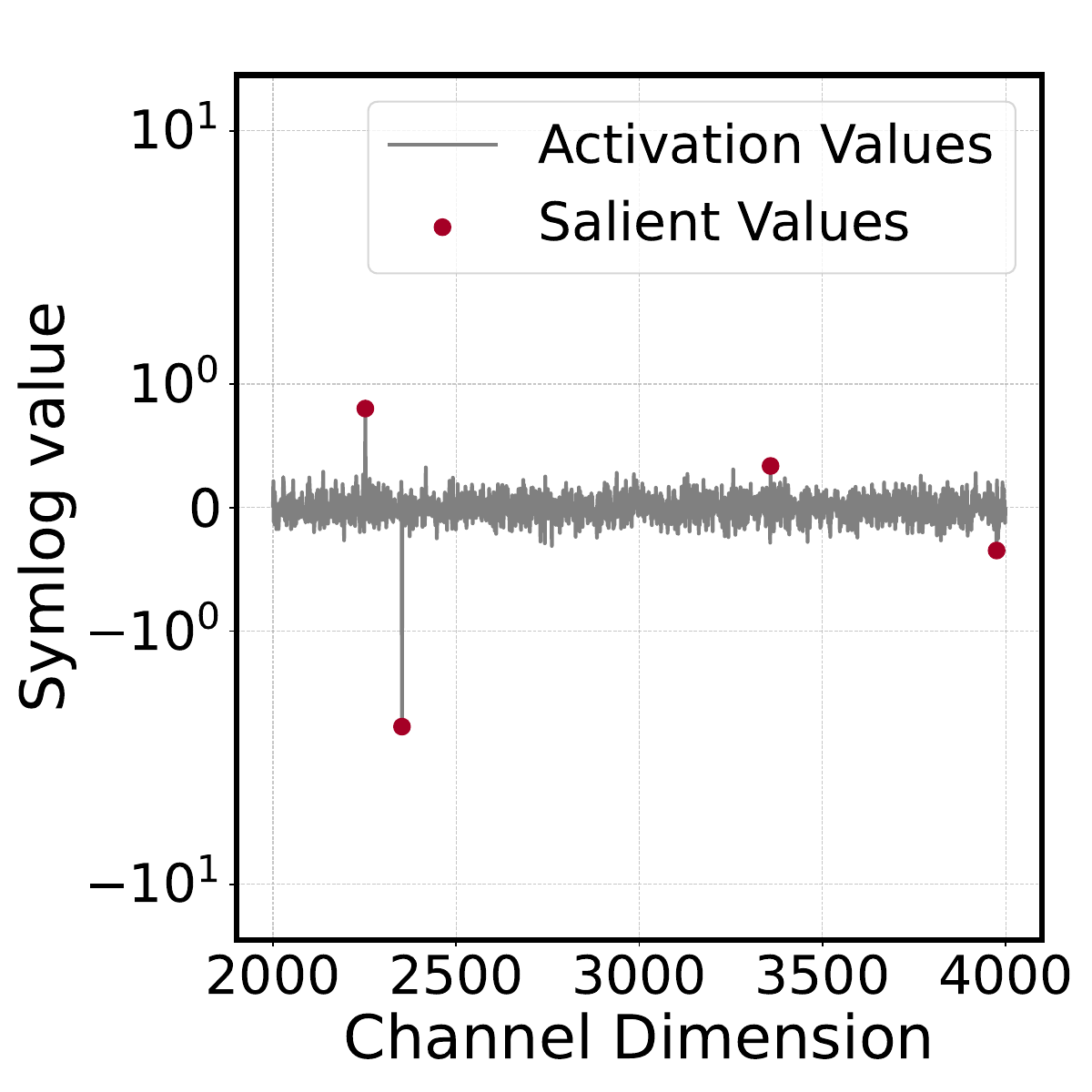}
    \caption{$down_{proj}$ of layer 26}
    \label{motivation2.sub.4}
\end{subfigure}

\caption{Activation distributions and salient values in Llama-3-8B under different module--layer combinations}
\label{motivation2}
\end{figure}



In weight–activation quantization, the presence of activation salient values (i.e., extreme-magnitude activations) constitutes a key factor contributing to the accuracy degradation of quantized large language models \cite{czako2025addressing}. To characterize this phenomenon more precisely, we analyze the activation statistics of Llama2-7B across all 32 transformer layers and seven representative linear modules ($gate_{proj}$, $down_{proj}$, $up_{proj}$, $k_{proj}$, $q_{proj}$, $v_{proj}$, $o_{proj}$) during a single forward pass with the input token “is”. Each activation tensor has shape $[B,S,H]$, where $B{=}1$, $S{=}1$, and $H{=}4096$. Owing to paper space limitation, for each module, we select activations from channel indices 2000–4000. Salient activations (marked in red) are detected using the Median Absolute Deviation (MAD) algorithm (detailed in Section~\ref{MAD-Guided Salient Activation Detection}). The complete layer-wise visualization is provided in Appendix~\ref{appendix:Activations Distribution}.

Figures~\ref{motivation2.sub.1} to \ref{motivation2.sub.4} illustrate the heterogeneity of activation distributions across layers and modules. The results reveal that salient activations are neither uniformly distributed nor temporally consistent. For example, even within the same module ($o_{proj}$), different layers exhibit distinct tail behaviors and outlier densities, indicating strong inter-layer variation in dynamic ranges. Likewise, within the same layer, the $q_{proj}$, $v_{proj}$, and $up_{proj}$ modules produce markedly different activation kurtosis and sparsity levels, reflecting module-specific quantization sensitivity. These observations highlight that activation statistics are inherently non-stationary and token-dependent, which complicates static quantization schemes that assume fixed or channel-wise invariant distributions.

\subsection{Challenges}
To address activation salient values in model quantization, a common strategy is to assign higher bitwidths to salient values through mixed-precision quantization \cite{dettmers2022gpt3, lin2024awq}. Despite its advantages in model compression, mixed-precision quantization during inference faces three key challenges:

\textbf{1) Dynamic distribution of activation salient values.} The distribution of salient values varies across layers, modules, and even individual tokens. Consequently, fixed-channel strategies fail to adequately capture all salient values, thereby limiting the accuracy of mixed-precision quantization and degrading model performance.

\textbf{2) Incompatibility of storage between salient and normal values.} Once activation salient values are quantized to high-bit, they cannot be stored together with low-bit normal values. This necessitates simultaneous reordering of both weight and activation channels, which incurs significant computational overhead.

\textbf{3) Increased overhead in linear transformations}. During linear operations, both activations and weights require costly dequantization procedures. This results in increased peak memory footprint and energy consumption, which may ultimately hinder or stall inference.

\begin{figure*}[t]
    \centering

    \begin{minipage}[t]{0.39\textwidth}
        \centering
        \includegraphics[height=5cm]{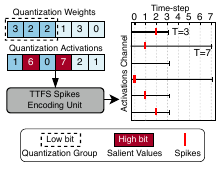}
        \captionof{figure}{Group weight-activation quantization and TTFS encoding}
        \label{workflow.sub.1}
    \end{minipage}
    \hfill
    \begin{minipage}[t]{0.6\textwidth}
        \centering
        \includegraphics[height=4.3cm]{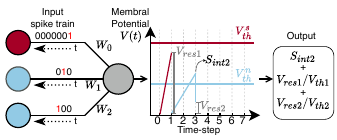}
        \captionof{figure}{Convertible SNN threshold: membrane potential accumulation and firing process}
        \label{workflow.sub.2}
    \end{minipage}
\end{figure*}

\section{SNN-based LLM Inference Framework Design}

To better handle salient values in ANN-based LLM quantization and address the above three challenges, we propose SpikeQuant. Specifically, we (i) use the offline MAD algorithm to obtain the salient bar for online accurate but low-cost detection of dynamic salient value, (ii) leverage TTFS encoding to convert quantized activations into binary spikes for mixed-precision storage, and (iii) integrate mixed-precision quantization scales into the SNN firing threshold, thereby avoiding both dequantization and costly dense MACs.

\subsection{MAD-Guided Salient Activation Detection}
\label{MAD-Guided Salient Activation Detection}

Different from prior works \cite{ardakani2022standard,park2018value}, we adopt the Median Absolute Deviation (MAD) \cite{rousseeuw1993alternatives}, a robust statistical estimator for outlier detection, to identify salient values along the hidden dimensions of each module’s activations during LLM inference. 

The use of the MAD  offers two key benefits. First, MAD is insensitive to extreme values, since the median (and deviations from it) is not strongly affected by outliers, unlike the standard deviation \cite{ardakani2022standard} which can be inflated by them. Second, unlike top-k methods \cite{park2018value} that rely on selecting a fixed number or fixed proportion of “largest” values, MAD-based thresholds adapt to the actual data distribution: when the number of salient values is unstable (varying across layers or inputs), a MAD-based criterion will still identify outliers accurately without the need to tune how many to pick.

For activations \(x\), MAD is defined as

\begin{equation}
    \mathrm{MAD} = \mathrm{median}\left( |x_i - \mathrm{median}(x)| \right),
\end{equation}
and the standardized score is computed as
\begin{equation}
    M_i = \frac{x_i - \mathrm{median}(x)}{c \cdot \mathrm{MAD}},
\end{equation}
where \(c\) is a scaling constant (typically \(c=1.4826\) \cite{iglewicz1993volume}) ensuring that the expected value of \(c \cdot \mathrm{MAD}\) to be roughly equal to the standard deviation for normally distributed data. Given a MAD threshold \(r\) (commonly \(r=3.5\) \cite{iglewicz1993volume}), any activation with \(|M_i| > r\) is regarded as a salient value. 

To avoid additional computational overhead during inference, we follow prior work \cite{shao2024omniquant,MLSYS2024_5edb57c0} and employ 128 randomly sampled sentences from the WikiText2 dataset as calibration data for offline salient value detection. Concretely, we store the average of the positive salient minimum and the negative salient maximum across 128 forward passes, referred to as the salient bar, which is then used as the criterion for online detection. This approach not only mitigates the limitations of fixed salient channels---which cannot capture all potential salient values---but also circumvents the expensive sorting operations required by top-\(k\) selection methods. Overall, it enables a lightweight yet accurate mechanism for dynamically identifying salient values during inference.

\subsection{TTFS Encoding of Mixed-Precision Activations}

After identifying salient values, we quantize them with a higher bitwidth (i.e. 5-bit) to preserve precision, while the majority of normal values are quantized with a lower bitwidth (i.e. 4-bit) using group-wise scaling \cite{yang2024gwq, park2018value}, thus better adapting the quantized range to local activation distributions. Specifically, in a single linear transformation module, activations have shape $[B,S,H]$ (batch, sequence, hidden) and weights have shape $[In,Out]$, producing outputs $[B,S,Out]$. Quantization groups ($G$ groups) are formed along the hidden dimension $H$ for activations and the input dimension $In$ for weights, while salient activation detection is performed along $H$, as illustrated in the left part of Fig. \ref{workflow.sub.1}

In practical hardware, however, mixed-bit activations cannot be directly stored in a single tensor due to constraints such as memory alignment requirements, SIMD/Systolic-array execution granularity, and the lack of heterogeneous-precision support in tensor cores \cite{rakka2024bf}. To resolve this limitation, we transform quantized activations into TTFS spike trains in the SNN domain. 

Formally, given a quantized activation value $Qa \in \{0,1,\ldots,2^b-1\}$ with bitwidth $b$, we define the encoding window length as $T = 2^b - 1$. Under TTFS encoding, the corresponding spike train is
\begin{equation}
    a(t) = 
    \begin{cases}
        1 & \text{if } t = T - Qa \\
        0 & \text{otherwise}
    \end{cases}
\label{eq8}
\end{equation}
Eq. (\ref{eq8}) means that the larger the $Qa$ value, the higher the membrane potential strength, and the earlier the spike occurs, as shown in Fig. \ref{workflow.sub.1}. This binary encoding method provides a different time window mapping across mixed precisions without requiring heterogeneous storage.

From a hardware perspective, spike generation is executed in parallel across activation channels. Each channel is equipped with a local spike generator, typically implemented using counters or stochastic firing units, which emits spike trains according to the precomputed firing rate. Because all activations are encoded within the same wall-clock window $T$, parallel spike streams are temporally aligned for subsequent synaptic operations. During computation, spikes serve as binary events that are multiplied with synaptic weights. This spike-weight multiplication is efficiently realized in neuromorphic accelerators via event-driven accumulate-and-shift operations, where each spike triggers the addition of the corresponding weight to the postsynaptic membrane potential. In this way, mixed-precision activations are seamlessly integrated into the SNN computation pipeline without requiring direct mixed-bit tensor storage.

\subsection{Dequantized ANN Equivalent SNN Inference}

After establishing the binary TTFS spike-based representation of activations, our objective is to ensure that the outputs of SNNs under mixed-precision quantization remain consistent with those obtained from the dequantized linear projections in ANN-based LLMs. Specifically, within a quantization group that contains only low-bit normal values, which contains no high-bit salient values, the dequantized linear transformation in a quantization group of the input sequence is expressed as

\begin{equation}
\begin{split}
    &Output_{ANN} = \sum_{i=1}^{H_g} (Qa_i - Za)\cdot Sa \cdot (Qw_i - Zw)\cdot Sw \\
&=\underbrace{ Sa \cdot Sw }_{z1} \cdot \underbrace{\sum_{i=1}^{H_g} \Big[ Qa_i \cdot (Qw_i - Zw) + Za \cdot (Zw - Qw_i) \Big] }_{z2} ,
\end{split}
\label{eq9}
\end{equation}
where $H_g$ denotes the quantization group size, $Qa_i$ and $Qw_i$ are the quantized activations and weights, respectively. $(Sa,Za)$ and $(Sw,Zw)$ represent the scaling factors and zero-points for activations and weights. Under symmetric quantization, the zero-points vanish.

From a mathematical perspective, the IF mechanism in SNNs can be interpreted as an accumulation of membrane potential $V_{acc}$, followed by emitting spikes whenever $V_{acc}$ exceeds the firing threshold $V_{th}$. In particular, the number of emitted spikes $S_{int}$ corresponds to the integer quotient of the membrane potential and the threshold:
\begin{equation}
    S_{int}\triangleq\left \lfloor V_{acc}\cdot \frac{1}{V_{th}} \right \rfloor. 
\label{eq10}
\end{equation}

Based on Eq.(\ref{eq9}), we observe that if we consider $Qa_i$ as TTFS spike train and $(Qw_i - Zw)$ as potential weight, the $z2$ term can be regarded as the accumulated potential $V_{acc}$ in SNNs:

\begin{equation}
V_{acc} = \underbrace{ \sum_{i=1}^{H_g}  (T-t_i)\cdot(Qw_i - Zw) }_{membrane\ potential\ V} + \underbrace{ \sum_{i=1}^{H_g} Za \cdot (Zw - Qw_i)}_{SNN\ bias\ b},
\label{eq11}
\end{equation}
where $t_i$ denotes the spike arrival time of element $i$, and $T$ is the encoding time window length of activation $Qa_i$. Based on $z1$ in Eq.(\ref{eq9}), Eq. (\ref{eq10}) and Eq. (\ref{eq11}), if we take $1/(Sa \cdot Sw)$ as $V_{th}$, then the fire spike count equals the integer part of the ANN output, which we denoted as $S_{int}$. The SNNs IF process is

\begin{equation}
    V(t) = V(t-1) + a(t)\cdot (T-t)\cdot(Qw_i-Zw)  +  b(t)- S(t) V_{\mathrm{th}}.
\label{eq12}
\end{equation}

\begin{equation}
S(t) =
\begin{cases}
0, & \text{if } V(t) < V_{\mathrm{th}}\\
1, & \text{if } V(t) \ge V_{\mathrm{th}}
\end{cases}.
\label{eq13}
\end{equation}

\begin{equation}
b(t) =
\begin{cases}
\sum_{i=1}^{H_g} Za \cdot (Zw - Qw_i), & \text{if } t=T\\
0, & \text{otherwise}
\end{cases}.
\label{eq14}
\end{equation}

Eq. (\ref{eq12}) shows that the membrane potential at time $t$ accumulates both the previous potential $V(t-1)$ and the contribution of the new spike. In TTFS coding, $a(t)\cdot (T-t)$ indicates that a spike at time $t$ corresponds to $(T-t)$ equivalent spikes, while $a(t)=0$ otherwise. The term $(Qw_i-Zw)$ represents the spike weight, and $-S(t)V_{\mathrm{th}}$ with Eq.~(\ref{eq13}) ensures that once the potential exceeds the threshold, a spike is emitted and $V(t)$ decreases by $V_{\mathrm{th}}$. Eq. (\ref{eq14}) further adds a bias term to the residual potential for a final firing. This bias $b(t)$ can be precomputed after quantization and stored before ANN-to-SNN conversion, incurring no runtime overhead.

Finally, the residual potential $V_{rest}$ that does not trigger a spike corresponds to a fractional part, denoted by $S_{float}$, yielding within each quantization group:

\begin{equation}
    S_{float} = \frac{V_{rest}}{V_{th}},
\end{equation}

\begin{equation}
output_{ANN}=output_{SNN} = S_{int} + S_{float}.
\end{equation}

This formulation establishes a principled connection between ANN dequantization and SNN rate-based computation. However, a key challenge arises because salient values are quantized with higher precision along the hidden dimension. Consequently, the scaling factor for salient values differs from that of each quantization group. To ensure equivalence with the ANN, the firing thresholds must also vary accordingly. After quantization, a 1-bit mask code is used to identify salient values. For each group, we pre-compute the normal threshold $V_{th}^n$ and the cross-group salient threshold $V_{th}^s$. This enables parallel IF operations within groups for normal values and across groups for salient values. The final SNN output for a sequence of activations and weights is given by Eq. \ref{eq17}, and the mix-precision membrane potential process is shown as Fig. \ref{workflow.sub.2}.

\begin{equation}
\text{Output}_{\text{SNN}} = (S_{int}^s + S_{float}^s )+ \sum_{g=1}^G \big( S_{int}^n(g) + S_{float}^n(g) \big).
\label{eq17}
\end{equation}

\begin{table*}[t]
  \caption{Zero-shot ($\uparrow$) performance of Llama2 and OPT models on  six common sense tasks. The best score is bolded.}
  \label{tab:quant_performance_full}
  \centering
  \newcolumntype{C}[1]{>{\centering\arraybackslash}p{#1}}
  
  \begin{tabular}{ll *{7}{C{1.1cm}}}
    \toprule
    Model & Method & PIQA & ARC\_e & ARC\_c & BoolQ & HellaSwag & Winogrande & Avg. \\
    \midrule
    
    \multirow{6}{*}{Opt-1.3B} 
    & Fp16 & 72.42 & 50.80 & 29.69 & 57.68 & 53.73 & 59.67 & 54.00 \\
    \cmidrule(l){2-9} 

    & GroupQuant & 67.79 & 42.93 & 25.26 & 49.51 & 45.94 & 54.93 & 47.73 \\
    & OneBit & 62.57 & 41.25 & 24.06 & \textbf{59.45} & 34.26 & 51.14 & 45.46 \\
    & LRQuant & 60.23 & 43.22 & 19.80 & 57.98 & 30.71 & 51.78 & 43.95 \\
    & GPTQ & 68.34 & 46.17 & \textbf{27.65} & 51.57 & 46.53 & 55.26 & 49.25 \\
    \rowcolor{gray!10}
    & SpikeQuant & \textbf{71.49} & \textbf{49.66} & 27.22 & 53.27 & \textbf{51.46} & \textbf{58.56} & \textbf{51.94} \\
    \midrule
    
    \multirow{6}{*}{Opt-2.7B} 
    & Fp16 & 74.81 & 54.34 & 31.31 & 60.28 & 60.59 & 60.93 & 57.04 \\
    \cmidrule(l){2-9}
    & GroupQuant & 72.91 & 49.96 & 29.44 & \textbf{62.26} & 55.94 & \textbf{59.75} & 55.04 \\
    & OneBit & 63.87 & 43.39 & 24.40 & 54.28 & 38.18 & 51.67 & 45.97 \\
    & LRQuant & 64.04 & 47.81 & 20.65 & 54.10 & 33.59 & 53.51 & 45.62 \\
    & GPTQ & 71.38 & 48.19 & 27.82 & 55.43 & 49.86 & 54.82 & 51.25 \\
    \rowcolor{gray!10}
    & SpikeQuant & \textbf{74.21} & \textbf{53.79} & \textbf{31.66} & 61.19 & \textbf{59.11} & \textbf{59.75} & \textbf{56.62} \\
    \midrule
    
    \multirow{7}{*}{Llama2-7B} 
    & Fp16 & 77.37 & 52.53 & 41.38 & 73.12 & 72.99 & 66.85 & 64.04 \\
    \cmidrule(l){2-9}
    & GroupQuant & 75.79 & 54.04 & \textbf{41.13} & 71.87 & 71.68 & 63.54 & 63.01 \\
    & SmoothQuant & 63.11 & 40.03 & 31.57 & 58.47 & 43.38 & 52.80 & 48.23 \\
    & OmniQuant & 66.15 & 45.20 & 31.14 & 63.51 & 56.44 & 53.43 & 52.65 \\
    & Atom & 76.28 & 52.10 & 38.99 & 69.79 & 69.81 & 63.69 & 61.78 \\
    & SpikeLLM & 64.47 & 48.74 & 27.30 & 63.27 & 43.29 & 56.83 & 50.65 \\
    \rowcolor{gray!10}
    & SpikeQuant & \textbf{77.09} & \textbf{55.39} & 40.36 & \textbf{73.49} & \textbf{72.09} & \textbf{65.27} & \textbf{63.95} \\
    \midrule
    
    \multirow{6}{*}{Llama2-13B} 
    & Fp16 & 79.05 & 59.85 & 44.62 & 68.53 & 76.22 & 70.09 & 66.39 \\
    \cmidrule(l){2-9}

    & GroupQuant &77.75 &55.64 &43.69 &66.45 &74.78 &68.3 &64.45 \\ 
    & SmoothQuant & 64.47 & 41.75 & 30.89 & 62.29 & 46.68 & 51.70 & 49.63 \\
    & OmniQuant & 69.69 & 47.39 & 33.10 & 62.84 & 58.96 & 55.80 & 54.63 \\
    & Atom & 77.69 & \textbf{57.58} & 42.92 & \textbf{67.46} & 73.77 & \textbf{68.51} & 64.66 \\
    & SpikeLLM & 66.49 & 55.30 & 30.12 & 64.16 & 47.43 & 51.54 & 52.51 \\
    \rowcolor{gray!10}
    & SpikeQuant &\textbf{79.33} &56.48 &\textbf{44.03} &66.15 &\textbf{75.53} &68.11 &\textbf{64.94} \\ 
    
    \bottomrule
  \end{tabular}
\label{tab:zero-shot}
\end{table*}

\section{Evaluation}

We conduct a comprehensive evaluation of SpikeQuant’s accuracy and energy efficiency. For accuracy, we report generation perplexity and zero-shot accuracy on standard benchmarks. For energy efficiency, we estimate the energy cost of MACs, accumulations, and data movement during inference. We also perform ablation studies to quantify the contribution of each component in SpikeQuant.

\subsection{Settings} 

\textbf{Training Details.} SpikeQuant employs 4-bit asymmetric quantization for both weights and activations without clipping, while salient values are quantized at 5 bits. For calibration, we randomly sample 128 segments of 2048 tokens from WikiText2 and use them to estimate saliency thresholds for each layer and module. Quantization is performed layer by layer, running on two NVIDIA A100 40GB GPUs.

\textbf{Baselines.} To comprehensively assess SpikeQuant’s accuracy performance, we compare against a diverse set of quantization baselines. For Llama2 models, we focus on methods that specifically address saliency in weights and activations to improve quantized accuracy. In particular, SmoothQuant \cite{xiao2023SmoothQuant} and OmniQuant \cite{shao2024omniquant} use 4-bit quantization for both weights and activations (W4A4), while Atom \cite{MLSYS2024_5edb57c0} and SpikeLLM \cite{xing2025spikellm} adopt W4A4 with salient values elevated to 8 bits. For OPT models, we include baselines spanning different weight quantization bitwidth: GroupQuant, OneBit \cite{xu2024onebit}, LRQuant \cite{zhao2024lrquant}, and GPTQ \cite{frantar2022gptq}. GroupQuant is a simple group-wise scheme with 128 groups. A complete description of baselines is in the Appendix \ref{appendix:Baseline details}.

\textbf{Evaluation Tasks.}
We evaluate SpikeQuant on widely used open-source LLaMA \cite{touvron2023llama} and OPT \cite{zhang2022opt} models. For model accuracy performance, we adopt standard metrics, including model accuracy, perplexity, and zero-shot accuracy. Language modeling perplexity is measured on the WikiText-2 \cite{merity2017pointer} and C4 \cite{raffel2020exploring} benchmarks. For zero-shot evaluation, we report accuracy on PIQA \cite{bisk2020piqa}, ARC-easy \cite{clark2018think}, ARC-challenge \cite{clark2018think}, BoolQ \cite{clark2019boolq}, HellaSwag \cite{clark2018think}, and Winogrande \cite{sakaguchi2021winogrande}. For inference energy-cost performance analysis, we use measured energy for compute, data movement, and memory access obtained from a commercial 22 nm process.

\subsection{Accuracy Evaluation}

\begin{table*}[t!]
  \newcolumntype{C}[1]{>{\centering\arraybackslash}p{#1}}
  
  \begin{subtable}[t]{0.49\textwidth}
    \centering
    \caption{Perplexity of Llama2 Models}
    \label{tab:llama2_models_booktabs}
    \begin{tabular}{ll C{1cm}C{0.8cm} C{1cm}C{0.8cm}}
      \toprule
      \multirow{2}{*}{Methods} & \multirow{2}{*}{Bits} & \multicolumn{2}{c}{Llama2-7B} & \multicolumn{2}{c}{Llama2-13B} \\
      \cmidrule(lr){3-4} \cmidrule(lr){5-6}
      & & Wikitext2 & C4 & Wikitext2 & C4 \\
      \midrule
      Fp16 & W16A16 & 5.68 & 7.08 & 4.88 & 6.46 \\
      \midrule
      SmoothQuant & W4A4 & 22.62 & 31.21 & 33.98 & 41.53 \\
      OmniQuant & W4A4 & 14.61 & 19.35 & 12.3 & 15.87 \\
      Atom & W4A(4\&8) & 6.16 & 7.70 & 5.46 & 7.03 \\
      SpikeLLM & W4A(4\&8) & 11.36 & 15.87 & 9.71 & 12.10 \\
      \rowcolor{gray!10}
      SpikeQuant & W4A(4\&5) & \textbf{5.79} & \textbf{7.33} & \textbf{5.04} & \textbf{6.65} \\
      \bottomrule
    \end{tabular}
  \end{subtable}
  \hfill 
  \begin{subtable}[t]{0.49\textwidth}
    \centering
    \caption{Perplexity of OPT Models}
    \label{tab:opt_models_booktabs}
    \begin{tabular}{ll C{1cm}C{0.8cm} C{1cm}C{0.8cm}}
      \toprule
      \multirow{2}{*}{Methods} & \multirow{2}{*}{Bits} & \multicolumn{2}{c}{Opt1.3B} & \multicolumn{2}{c}{Opt2.7B} \\
      \cmidrule(lr){3-4} \cmidrule(lr){5-6}
      & & Wikitext2 & C4 & Wikitext2 & C4 \\
      \midrule
      Fp16 & W16A16 & 14.63 & 14.72 & 12.47 & 13.17 \\
      \midrule
      GroupQuant & W4A4 & 36.03 & 31.08 & 15.19 & 15.64 \\
      OneBit & W1A16 & 25.42 & 22.95 & 21.86 & 20.76 \\
      LRQuant & W2A16 & 48.43 & 69.34 & 30.59 & 42.49 \\
      GPTQ & W3A16 & 20.97 & 21.63 & 16.88 & 18.17 \\
      \rowcolor{gray!10}
      SpikeQuant & W4A(4\&5) & \textbf{16.38} & \textbf{16.56} & \textbf{13.15} & \textbf{13.91} \\
      \bottomrule
    \end{tabular}
  \end{subtable}
  
  \caption{Perplexity ($\downarrow$) performance of Llama2 and OPT models on Wikitext2 and C4 datasets. The best score is bolded.}
  \label{tab:perplexity}
\end{table*}

\begin{table*}[t]
  \caption{Energy Ratio of Different Computing Units and Data Movement Operations (Normalized to 4-4-4 ACC).}
  \label{tab:energy_computing_data_movement}
  \centering
  \newcolumntype{C}[1]{>{\centering\arraybackslash}p{#1}}
  \begin{tabular}{C{1.2cm} *{8}{C{1.1cm}} *{2}{C{1.1cm}}}
    \toprule
    & \multicolumn{8}{c}{Computing} & \multicolumn{2}{c}{Data movement} \\
    \cmidrule(lr){2-9}\cmidrule(lr){10-11}
    Process 
      & 4-4-4 ACC 
      & 5-5-5 ACC 
      & 4-4-32 MAC 
      & 4-5-32 MAC
      & 4-8-32 MAC 
      & 1-16-32 MAC 
      & 2-16-32 MAC 
      & 3-16-32 MAC 
      & Read/bit 
      & Move/bit \\
    \midrule
    Energy 
       & 1.00 & 1.18 & 8.66 & 9.24 & 10.94 & 10.89 & 11.46 & 13.28 & 6.04 &11.04 \\
    \bottomrule
  \end{tabular}
\label{tab:energy_value}
\end{table*}

\begin{figure*}[t] 
  \centering
  \includegraphics[width=\textwidth]{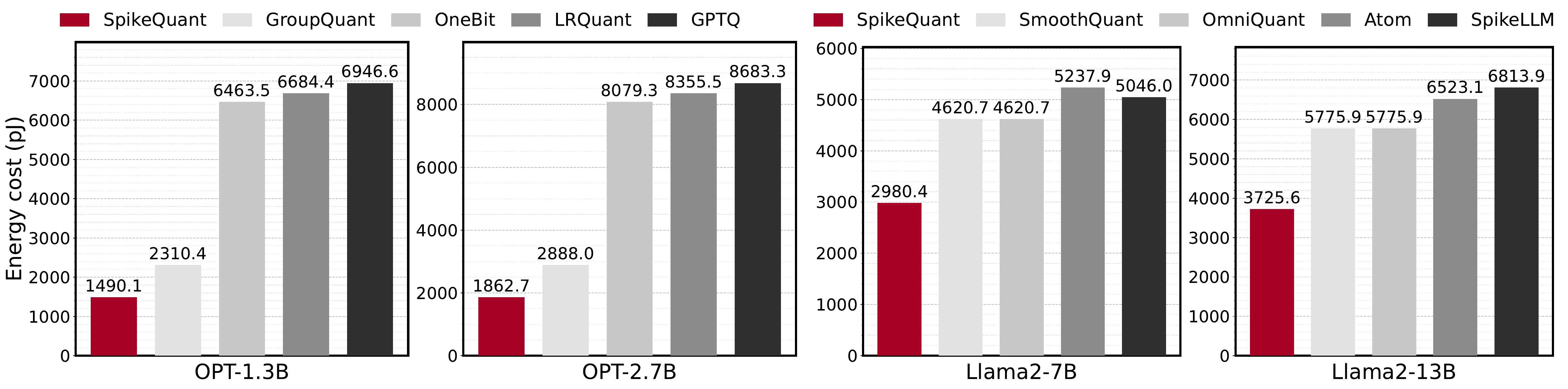}
  \caption{Energy cost comparison across OPT and Llama models.}
  \label{fig:energy_comparision}
\end{figure*}

As shown in Table \ref{tab:zero-shot}, we compare zero-shot accuracy on six tasks across Llama and OPT models. SpikeQuant consistently outperforms other weight–activation quantization baselines. Among salient aware W4A4 methods, SpikeQuant incurs only 0.09\% and 1.45\% average accuracy loss on Llama-7B and Llama-13B versus FP16, whereas prior work reports up to 15.81\%  and 16.76\% on Llama-7B and 13B under the same settings. For baselines that reduce weight bit-width to improve compression, SpikeQuant likewise maintains low precision with only 2.60\% (OPT-1.3B) and 0.42\% (OPT-2.7B) accuracy loss, compared to losses as high as 10.05\% and 11.42\% for similar methods.

Table \ref{tab:perplexity} reports perplexity on Llama and OPT. SpikeQuant further lowers perplexity, with 0.11 decrease on WikiText-2 for Llama-7B. On OPT, it largely preserves perplexity, averaging a 0.71 decrease on OPT-2.7B, while existing methods fail to achieve comparable levels.

SpikeQuant’s strong accuracy and perplexity stem from precisely identifying salient values and quantizing them in place at higher bit-width. In contrast, Atom and SpikeLLM, although elevating salient values to 8-bit, rely on fixed salient channels or fixed salient ratios, which fail to accurately localize saliency. This can inflate the quantization scale spanning normal activations and salient values, reducing quantization fidelity and ultimately lowering model accuracy. Additional experiments on the number and proportion of salient values are in the Appendix \ref{appendix:Salient Value Ratio and Counts}.

\subsection{Energy Efficiency Evaluation}

Followed by prior works \cite{yan2024reconsidering, yan2025otters}, the inference energy in digital circuits comprises computation energy ($E_{\text{compute}}$) and data movement energy ($E_{\text{data}}$):

\begin{equation}
    E=E_{\text{compute}}+E_{\text{data}}.
\label{eq18}
\end{equation}
The computation energy $E_{\text{compute}}$ accounts for arithmetic operations of MACs and ACCs. The data movement energy $E_{\text{data}}$ covers the energy to read weights from SRAM and to move data from previous-layer outputs to the current layer’s computation.

Our baselines fall into three categories: typical weight–activation quantization (e.g., SmoothQuant and OneBit), mixed-precision weight –activation quantization (e.g., Atom), and SNN-based mixed-precision weight–activation quantization (e.g., Spike LLM). Our energy model assumes a spatial dataflow architecture with a Network-on-Chip (NoC), a design common to neuromorphic hardware like Loihi~\cite{lines2018loihi} and TrueNorth~\cite{akopyan2015truenorth} and large-scale AI accelerators such as Sambanova~\cite{prabhakar2022sambanova} and Cerebras~\cite{lie2022cerebras}. Based on the dataflow architecture and Eq. (\ref{eq18}), we derive energy-cost formulas for these three baseline categories (provided in detail in the Appendix \ref{appendix:Energy Cost Formulation}). To evaluate hardware costs, we synthesized MAC and ACC units at the bit widths used by each baseline in a commercial 22 nm process. Table \ref{tab:energy_value} reports the resulting energy measurements.

Using the measurements in Table \ref{tab:energy_value} and the energy formulas, we obtain the final energy-cost comparison results shown in Figure \ref{fig:energy_comparision}. The results show that SpikeQuant, while achieving higher accuracy, reduces energy by up to 78.5\% (OPT-2.7B, compared to GPTQ). The savings primarily stem from SpikeQuant’s lower quantization bit widths and the fact that, for each activation parameter, only a 1-bit TTFS spike encoding needs to be moved.

\begin{table*}[t]
  \caption{Perplexity ($\downarrow$) ablation study on different quantization techniques used in SpikeQuant.}
  \label{tab:ppl_opt_llama_avg_only}
  \centering
  \newcolumntype{C}[1]{>{\centering\arraybackslash}p{#1}}
  \begin{tabular}{l C{2.8cm} C{2.8cm} C{2.2cm} C{2.2cm}}
    \toprule
    Method & OPT-1.3B & OPT-2.7B & Llama2-7B & Llama2-13B \\
    \midrule
    FP16 
      & 14.68 & 12.82 & 6.38 & 5.67 \\
    \midrule
    + W4A4 
      & 33402.11 (+33387.43) & 12668.00 (+12655.18) & 660.33 (+653.95) & 411.64 (+405.97) \\
    + Group size 256 
      & 259.10 (-33143.01) & 83.12 (-12584.88) & 7.16 (-653.17) & 6.29 (-405.35) \\
    Group size 128 
      & 33.55 (-225.55) & 15.41 (-67.71) & 6.84 (-0.32) & 6.02 (-0.27) \\
    + Fix 128 salient channel 
      & 35.10 (+1.55) & 17.28 (+1.87) & 6.93 (+0.09) & 6.25 (+0.23) \\
    + SpikeQuant (MAD $r=5$) 
      & 16.54 (-18.56) & 13.55 (-3.73) & 6.60 (-0.33) & 5.86 (-0.39) \\
    SpikeQuant (MAD $r=3.5$) 
      & 16.47 (-0.07) & 13.53 (-0.02) & 6.56 (-0.04) & 5.85 (-0.01) \\
    \bottomrule
  \end{tabular}
\label{tab:ablation}
\end{table*}

\subsection{Ablation Study}

Table \ref{tab:ablation} presents the perplexity results of our ablation study on various components in SpikeQuant. As shown, the baseline FP16 models achieve low perplexity across all scales. However, direct W4A4 quantization causes severe performance degradation, with perplexity increasing by several orders of magnitude, indicating that naive low-bit quantization leads to substantial information loss. Introducing group-wise quantization (group size = 256)  mitigates this issue, reducing perplexity to a reasonable level. Further decreasing the group size to 128 yields additional improvements, demonstrating the benefit of finer granularity. When we fix the top 128 salient channels, the perplexity does not improve, suggesting that salient activations vary dynamically across forward passes and cannot be captured by static channel selection.

In contrast, SpikeQuant adaptively identifies salient channels at runtime using a MAD-based criterion, allowing flexible precision allocation without introducing instability. This adaptive mechanism enables SpikeQuant to achieve significantly lower perplexity than all static quantization variants. Notably, even when the baseline perplexity is already close to the FP16 level, SpikeQuant continues to yield measurable reductions (e.g., from 35.10 to 16.47 for OPT-1.3B and from 6.25 to 5.85 for Llama2-13B)

\section{Related Works}

\textbf{Model quantization.} Quantization reduces model size by converting floating-point parameters to low-bit representations, typically down to 2, 4, or 8 bits \cite{chowdhury2021spatio, deng2021comprehensive, sorbaro2020optimizing} , enabling efficient deployment on resource-constrained devices \cite{gholami2022survey,liang2021pruning}. The progression from weight quantization \cite{kashiwamura2024effect, ajit2024cdquant} to weight-activation quantization \cite{gai2024atq, lee2023enhancing} has significantly advanced neural network compression. However, prior works observed that a key challenge in LLM quantization is the salient values (also called outliers) phenomenon \cite{an2025systematic, linAWQ2024}. These large-magnitude values are challenging as they distort the quantization range, leading to significant precision loss for other values\cite{ lin2024rotation}. Most existing methods mitigate this by focusing on fixed salient channels in weights or activations.\cite{MLSYS2024_5edb57c0, wei-etal-2023-outlier, nrusimha2024mitigating} This approach, however, overlooks that activation distributions are dynamic and vary with each forward pass during inference.

\textbf{Spike neural networks.} Spiking Neural Networks (SNNs) \cite{roy2019towards, maass1997networks} emerge as a promising brain-inspired computational model that mimics the binary spike information transmission mechanism of biological neurons, drawing from fundamental neuroscience principles of neural communication through action potentials \cite{mehonic2022brain, zhang2020system}.  Current SNNs research predominantly focuses on computer vision tasks \cite{wei2024q, hong2025lasnn} and small-scale language 
processing tasks \cite{bal2024spikingbert, zhu2023spikegpt} by converting ANNs to SNNs, where SNNs offer energy-efficient alternatives to traditional ANNs due to their high biological plausibility. However, applications of SNNs to LLMs remain significantly underexplored, representing a substantial gap in current research directions despite the potential for energy-efficient language processing that could leverage the temporal dynamics inherent in spiking mechanisms.

\section{Conclusion}

The rapid growth of ANN-based LLMs has driven new demands for efficient deployment in Web-of-Things (WoT) environments, where privacy, latency, and energy constraints are critical. To address these challenges, we proposed SpikeQuant, a brain-inspired quantization framework that integrates spiking dynamics into LLM computation. By encoding salient activations as spike counts and embedding quantization scales into firing thresholds, SpikeQuant enables dequantization-free and energy-efficient inference. Experiments show that it preserves near-FP16 accuracy under W4A4 quantization and further lowers perplexity when baseline performance is already high, demonstrating its potential for low-power, high-accuracy LLM deployment in WoT scenarios.



\bibliographystyle{unsrt}
\bibliography{main}

\appendix

\section{Baseline details}
\label{appendix:Baseline details}

\textbf{SmoothQuant} \cite{xiao2023SmoothQuant} is a post-training quantization (PTQ) method that transfers part of the quantization “difficulty” from activations to weights. It rescales activation channels and applies corresponding compensation to the weights, making activations easier to represent at low bitwidth. This approach enables W8A8 (8-bit weight and activation) quantization while maintaining minimal accuracy loss on large language models.

\textbf{OmniQuant} \cite{shao2024omniquant} is a differentiable and learnable quantization framework featuring two key modules: Learnable Weight Clipping, which regulates extreme weight values, and Learnable Equivalent Transformation, which shifts activation outliers into the weight domain. By minimizing quantization error at the block level, OmniQuant achieves strong performance even under extremely low-bit settings such as W4A4 or W3A16.

\textbf{Atom} \cite{MLSYS2024_5edb57c0} targets service-oriented, throughput-optimized low-bit deployment. It combines mixed-precision, fine-grained group quantization, dynamic activation quantization, and KV-cache quantization, enabling efficient 4-bit weight and activation inference. Atom achieves substantial speedups over FP16 and INT8 while maintaining competitive accuracy.

\textbf{SpikeLLM} \cite{xing2025spikellm} integrates Spiking Neural Network (SNN) principles with quantization and discretization mechanisms for language modeling. It introduces an “Optimal Brain Spiking” quantization framework that selectively applies spiking or quantization to specific channels, reducing energy consumption and accelerating inference. Moreover, SpikeLLM is compatible with existing quantization pipelines such as GPTQ and OmniQuant.

\textbf{OneBit} \cite{xu2024onebit} explores extreme low-bit quantization, compressing LLM weight matrices to 1 bit (or near-binary precision). It designs specialized 1-bit parameter representations and matrix-factorization-based initialization to improve convergence after quantization. Experiments show that OneBit can achieve reasonable performance retention on models such as LLaMA under ultra-low bitwidths.

\textbf{LRQuant} \cite{zhao2024lrquant} is a learnable and robust PTQ approach built upon SmoothQuant and OmniQuant. It introduces a new NLC + MSE loss and a learnable smoothing-parameter mechanism. Additionally, it proposes Gamma Migration, which shifts the LayerNorm $\gamma$ parameter into subsequent layer weights or residual paths, thereby stabilizing activation outlier distributions and improving quantization robustness.

\textbf{GPTQ} \cite{frantar2022gptq} is a one-shot weight quantization method tailored for GPT- and OPT-style generative transformers. It leverages approximate second-order statistics (e.g., Hessian or inverse-variance approximations) to guide quantization, maintaining low error even at 3–4 bit precision. OPTQ enables large models to be quantized within a few GPU hours with negligible accuracy degradation.

\begin{table*}[t]
    \centering
    \caption{Ratio and number of salient values comparison in OPT and Llama2 models.}
    \label{tab:opt13b_saliency}
    \setlength{\tabcolsep}{6pt}
    \begin{tabular}{lrrrrrrrr}
        \toprule
        Methods &
        \multicolumn{2}{c}{OPT1.3B (H=2048)} &
        \multicolumn{2}{c}{OPT1.3B (H=2560)} &
        \multicolumn{2}{c}{OPT1.3B (H=4096)} &
        \multicolumn{2}{c}{OPT1.3B (H=5120)} \\
        \cmidrule(lr){2-3} \cmidrule(lr){4-5} \cmidrule(lr){6-7} \cmidrule(lr){8-9}
        & Ratio & Number
        & Ratio & Number
        & Ratio & Number
        & Ratio & Number \\
        \midrule
        SpikeQuant & 2.45\% & 50  & 2.15\% & 55  & 2.68\% & 110 & 2.30\% & 118 \\
        Atom       & 6.25\% & 128 & 5.00\% & 128 & 3.12\% & 128 & 2.50\% & 128 \\
        SpikeLLM   & 10.00\% & 205 & 10.00\% & 256 & 10.00\% & 410 & 10.00\% & 512 \\
        \bottomrule
    \end{tabular}
\label{tab:salient_ratio}
\end{table*}

\section{Energy Cost Formulation}
\label{appendix:Energy Cost Formulation}

We model the energy cost of a linear transformation within one Transformer block under different quantization and encoding schemes. The following notations are used throughout:
\begin{itemize}
  \item $B$, $S$, $H$, and $\mathrm{Out}$ denote the batch size, sequence length, hidden dimension, and output dimension, respectively.
  \item $b_w$ and $b_a$ represent the bitwidths of weights and activations.
  \item $E_{\mathrm{MAC}}$ and $E_{\mathrm{ACC}}$ are the per-operation energy costs of a multiply--accumulate (MAC) and accumulate (ACC) operation.
  \item $E_{\mathrm{read}}$ and $E_{\mathrm{move}}$ denote the energy costs of memory read and data movement.
  \item $\gamma$ indicates the number of salient activation channels within a hidden group.
  \item $T^{\mathrm{high}}$ and $T^{\mathrm{low}}$ are the spike time window for salient and non-salient activations, and $S_r^{\mathrm{high}}$, $S_r^{\mathrm{low}}$ represent their spike rates.
\end{itemize}

For typical quantization, the computation term scales with the total number of MAC operations, while the data movement term reflects memory access proportional to weight and activation bitwidths. The energy cost of a linear transform in one block is:

\begin{equation}
\begin{split}
        E_q = &\underbrace{B\cdot S\cdot Out\cdot  H\cdot E_{MAC}}_{E_{\text{compute}}} \\
    &+\underbrace{In\cdot Out\cdot b_w \cdot E_{read} +  B\cdot S\cdot H\cdot b_a \cdot E_{move}}_{E_{\text{data}}}
\end{split}
\end{equation}

For mix-precision quantization with $\gamma$ salient activation values in a hidden group, we partitions the energy into high-bit and low-bit computation paths. The high-bit path handles salient activations, consuming more MAC energy and data bandwidth, while the low-bit path handles the rest. The total energy of a linear transform in one block is their sum:
\begin{equation}
    E_m = E_q^{high}+E_q^{low}
\end{equation}

\begin{equation}
\begin{split}
        E_q^{high} = &\underbrace{B\cdot S\cdot Out\cdot \gamma \cdot E_{MAC}^{high}}_{E_{compute}} \\
    &+\underbrace{\gamma\cdot Out\cdot b_w \cdot E_{read} +  B\cdot S\cdot \gamma\cdot b_a^{high} \cdot E_{move}}_{E_{data}}
\end{split}
\end{equation}

\begin{equation}
\begin{split}
        E_q^{low} = &\underbrace{B\cdot S\cdot Out\cdot (H-\gamma) \cdot E_{MAC}^{low}}_{E_{compute}} \\
    &+\underbrace{(H-\gamma)\cdot Out\cdot b_w \cdot E_{read} +  B\cdot S\cdot (H-\gamma)\cdot b_a^{low} \cdot E_{move}}_{E_{data}}
\end{split}
\end{equation}

For TTFS encoding based SpikeQuant with $\gamma$ salient activation values in a hidden group, the energy cost of a linear transform in one block is

\begin{equation}
    E_s = E_s^{high}+E_s^{low}
\end{equation}

\begin{equation}
\begin{split}
    E_s^{high} =&\underbrace{B\cdot S\cdot Out \cdot \gamma \cdot T^{high} \cdot S_r^{high}\cdot \big (E_{ACC}^{high}+E_{MAC}^{high}\big)}_{E_{compute}} \\
    &+\underbrace{\gamma\cdot Out\cdot b_w \cdot E_{read} +  B\cdot S\cdot \gamma \cdot T^{high} \cdot S_r^{high} \cdot E_{move}}_{E_{data}}
\end{split}
\end{equation}

\begin{equation}
\begin{split}
    E_s^{low} &=\underbrace{B\cdot S\cdot  Out \cdot (H-\gamma) \cdot T^{low} \cdot S_r^{low}\cdot \big (E_{ACC}^{low}+E_{MAC}^{low}\big)}_{E_{compute}} \\
    &+\underbrace{(H-\gamma)\cdot Out\cdot b_w \cdot E_{read} +  B\cdot S \cdot (H-\gamma)\cdot T^{low} \cdot S_r^{low} \cdot E_{move}}_{E_{data}}
\end{split}
\end{equation}

\section{Additional Experiments Results}

\subsection{Salient Value Ratio and Counts}
\label{appendix:Salient Value Ratio and Counts}

Table~\ref{tab:salient_ratio} compares the ratio and number of salient activation values identified by different methods with varying hidden dimensions. As shown, SpikeQuant consistently detects a much smaller proportion of salient activations (around 2–3\%) compared to Atom (2.5–6.25\%) and SpikeLLM (10\%). This indicates that SpikeQuant achieves more precise and selective saliency detection, effectively distinguishing truly critical activation channels while maintaining the high accuracy.

\subsection{Activations Distribution}
\label{appendix:Activations Distribution}

\begin{figure*}[t]
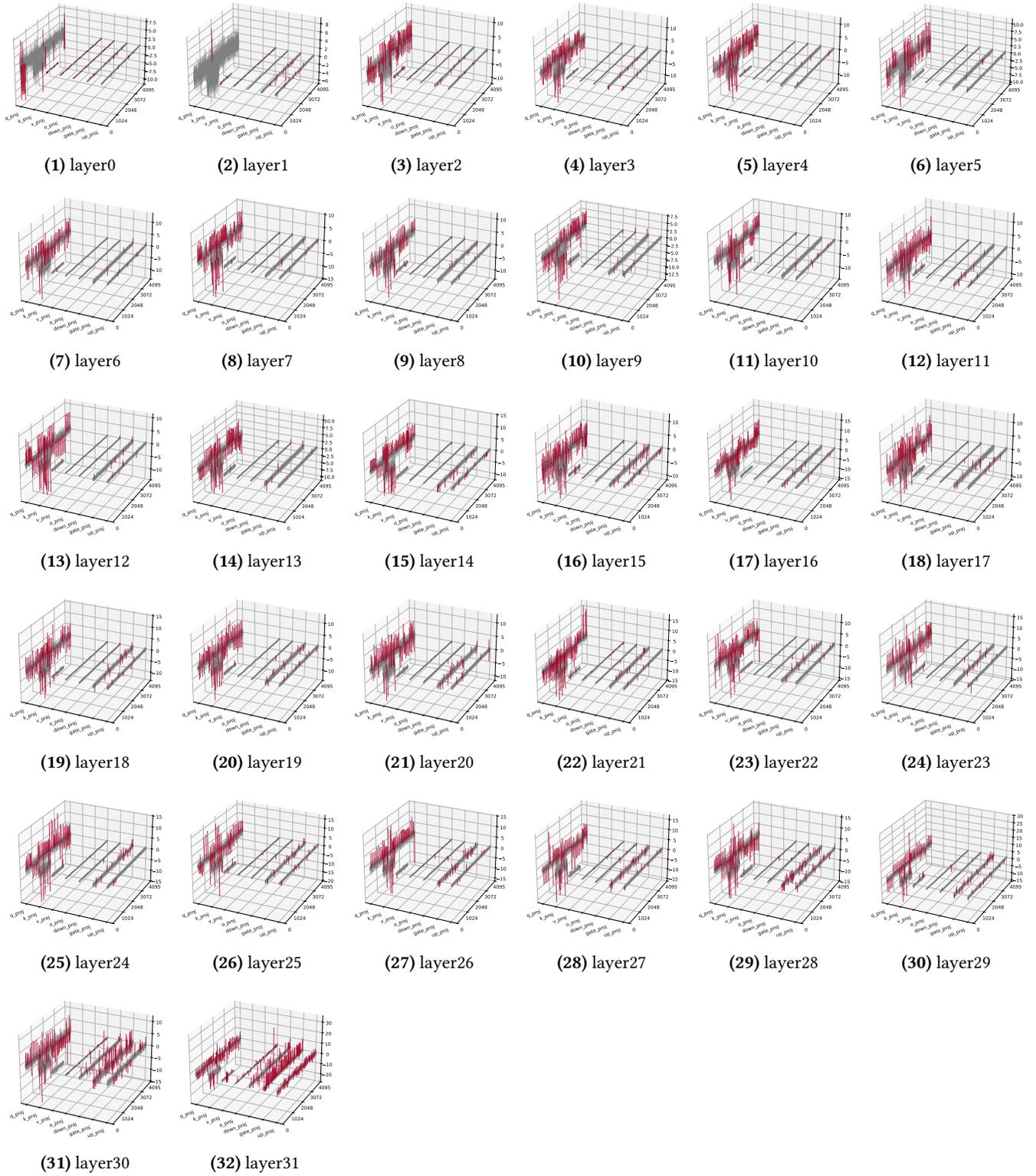

    \centering
    \foreach \i in {0,...,5}{
        \begin{subfigure}{0.155\linewidth}
            \centering
            \includegraphics[width=\linewidth]{./pics/exp/inference_3d/layer_\i.pdf}
            \caption{layer\i}
        \end{subfigure}\hfill
    }
    \vspace{0.6ex}
    \foreach \i in {6,...,11}{
        \begin{subfigure}{0.155\linewidth}
            \centering
            \includegraphics[width=\linewidth]{./pics/exp/inference_3d/layer_\i.pdf}
            \caption{layer\i}
        \end{subfigure}\hfill
    }
    \vspace{0.6ex}
    \foreach \i in {12,...,17}{
        \begin{subfigure}{0.155\linewidth}
            \centering
            \includegraphics[width=\linewidth]{./pics/exp/inference_3d/layer_\i.pdf}
            \caption{layer\i}
        \end{subfigure}\hfill
    }
    \vspace{0.6ex}
    \foreach \i in {18,...,23}{
        \begin{subfigure}{0.155\linewidth}
            \centering
            \includegraphics[width=\linewidth]{./pics/exp/inference_3d/layer_\i.pdf}
            \caption{layer\i}
        \end{subfigure}\hfill
    }
    \vspace{0.6ex}
    \foreach \i in {24,...,29}{
        \begin{subfigure}{0.155\linewidth}
            \centering
            \includegraphics[width=\linewidth]{./pics/exp/inference_3d/layer_\i.pdf}
            \caption{layer\i}
        \end{subfigure}\hfill
    }
    \vspace{0.6ex}
    \foreach \i in {30,...,31}{
        \begin{subfigure}{0.155\linewidth}
            \centering
            \includegraphics[width=\linewidth]{./pics/exp/inference_3d/layer_\i.pdf}
            \caption{layer\i}
        \end{subfigure}\hfill
    }
    \foreach \dummy in {1,...,4}{
        \begin{subfigure}{0.155\linewidth}
        \phantom{\rule{\linewidth}{0.75\linewidth}}
        \end{subfigure}\hfill
    }
    \caption{Activation distribution and salient values distribution on Llama2-7B.}
\label{fig:all_activation}
\end{figure*}

Figure \ref{fig:all_activation} illustrates the activation distribution of Llama2-7B during inference on the token “is”. The red lines indicate salient activations identified by the MAD-based algorithm, while the gray ones represent normal activations. From the results, we observe that even within a single forward pass for the same token, the distribution and number of salient values vary across layers and modules. This further demonstrates the rationality and flexibility of SpikeQuant in dynamically determining both the quantity and locations of salient values.

\end{document}